\documentclass{article}

\PassOptionsToPackage{numbers, compress}{natbib}
\usepackage[final]{nips}

\usepackage[utf8]{inputenc} 
\usepackage[T1]{fontenc}    
\usepackage{hyperref}       
\usepackage{url}            
\usepackage{booktabs}       
\usepackage{amsfonts}       
\usepackage{nicefrac}       
\usepackage{microtype}      

\usepackage{epsfig}
\usepackage{graphicx}
\usepackage{amsmath}
\usepackage{amssymb}
\usepackage{algorithm,algorithmic}
\usepackage{subfigure} 
\usepackage{multirow}
\usepackage{hyphenat}

\title{Multipartite Pooling \\ for Deep Convolutional Neural Networks}

\author{
  Arash shahriari \\
  Research School of Engineering \\
  Australian National University \\
  Canberra, Australia \\
  \texttt{arash.shahriari@anu.edu.au} \\
  \And
  Fatih Porikli \\
  Research School of Engineering \\
  Australian National University \\
  Canberra, Australia \\
  \texttt{fatih.porikli@anu.edu.au} \\
}

\begin{document}

\maketitle

\begin{abstract}
\nohyphens{We propose a novel pooling strategy that learns how to adaptively rank deep convolutional features for selecting more informative representations. To this end, we exploit discriminative analysis to project the features onto a space spanned by the number of classes in the dataset under study. This maps the notion of labels in the feature space into instances in the projected space. We employ these projected distances as a measure to rank the existing features with respect to their specific discriminant power for each individual class. We then apply multipartite ranking to score the separability of the instances and aggregate one-versus-all scores to compute an overall distinction score for each feature. For the pooling, we pick features with the highest scores in a pooling window instead of maximum, average or stochastic random assignments. Our experiments on various benchmarks confirm that the proposed strategy of multipartite pooling is highly beneficial to consistently improve the performance of deep convolutional networks via better generalization of the trained models for the test-time data.}
\end{abstract}

\section{Introduction}

The considerable complexity of object recognition makes it an interesting research topic in computer vision. Deep neural networks have recently addressed this challenge, with close precision to human observers. They recognize thousands of objects from millions of images, by using the models with large learning capacity. This paper proposes a novel pooling strategy that learns how to rank convolutional features adaptively, allowing the selection of more informative representations. 

To this end, the Fisher discrimination is exploited, to project the features into a space, spanned by the number of classes in the dataset under study. This mapping is employed as a measure to rank the existing features, with respect to their specific discriminant power, for each classes. Then, multipartite ranking is applied to score the separability of instances, and to aggregate one-versus-all scores, giving an overall distinction score for each features. For the pooling, features with the highest scores are picked in a pooling window, instead of maximum, average or stochastic random assignments. 

Spatial pooling of convolutional features, is critical in many deep neural networks. Pooling aims to select and aggregate features over a local reception field, into a local bag of representations, that are compact and resilient to transformations and distortions of the input~\cite{boureau2010theoretical}. Common pooling strategies often take sum~\cite{fukushima1988neocognitron}, average~\cite{le1990handwritten}, or maximum~\cite{jarrett2009best} response. There are also variants that enhance maximum pooling performance, such as, generalized maximum pooling~\cite{murray2014generalized} or fractal maximum pooling~\cite{graham2014fractional}. Deterministic pooling can be extended to stochastic alternatives, e.g. random selection of an activation within the pooling region, according to a multinomial distribution~\cite{zeiler2013stochastic}.

\section{Method}

There exists a vast literature on instance selection and feature ranking. Instance selection regimes usually belong to either condensation or edition proposals~\cite{leyva2015three}. They attempt to find a subset of data, in which, a trained classifier is provided with, similar or close validation error, as the primary data. Condensed Nearest Neighbour~\cite{hilborn1968dg}, searches for a consistent subset, where every instance inside is assumed to be correctly classified. Some variants of this method are Reduced Nearest Neighbour~\cite{gowda1979condensed}, Selective Nearest Neighbour~\cite{ritter1975algorithm}, Minimal Consistent Set~\cite{dasarathy1994minimal}, Fast Nearest Neighbour Condensation~\cite{angiulli2007fast}, and Prototype Selection by Clustering~\cite{olvera2010new}. 

In contrast, Edited Nearest Neighbour~\cite{wilson1972asymptotic}, discards the instances that disagree with the classification responses of their neighbouring instances. Some revisions of this strategy, are Repeated Edited Nearest Neighbour~\cite{tomek1976experiment}, Nearest Centroid Neighbour Edition~\cite{sanchez2003analysis}, Edited Normalized Radial Basis Function~\cite{jankowski2004comparison}, and Edited Nearest Neighbour Estimating Class Probabilistic and Threshold~\cite{vazquez2005stochastic}.

On the other hand, the family of feature ranking algorithms can be mainly grouped into, preference learning, bipartite, multipartite, or multilabel ranking. In situations where the instances have only binary labels, the ranking is called bipartite. Different aspects of bipartite ranking have been investigated in numerous studies including, RankBoost~\cite{freund2003efficient}, RankNet~\cite{burges2005learning}, and AUC maximizing~\cite{brefeld2005auc}, which are the ranking versions for AdaBoost, logistic regression, and Support Vector Machines. 

There are also several ranking measures, such as, average precision and Normalized Discounted Cumulative Gain. For multilabel instances, multipartite ranking approaches seek to maximize the volume under the ROC surface~\cite{waegeman2011era}, which is in contrast with the minimization of the pairwise ranking cost~\cite{uematsu2015statistical}.

The problem of employing either instance selection or feature ranking methods for pooling in deep neural networks, appears at the testing phase of trained models. The existing ranking algorithms, mostly deals with the training-time ranking. As a result, they are not usually advantageous for the pooling of convolutional features in the test phase. Without pooling, the performance of deep learning architectures degrades substantially. The local feature responses propagate less effectively to neighboring receptive fields, thus the local-global representation power of the convolutional network diminishes. Moreover, the network becomes very sensitive to input deformations. 

To tackle the above issues, a novel strategy \textit{i.e.} multipartite pooling, is introduced in this paper. This ranks convolutional features by employing supervised learning. In supervised learning, the trained scoring function reflects the ordinal relation among class labels. The multipartite pooling scheme learns a projection from the training set. Intuitively, this is a feature selection operator, whose aim is to pick the most informative convolutional features, by learning a multipartite ranking scheme from the training set. Inspired by stochastic pooling, higher ranked activations in each window, are picked with respect to their scoring function responses. Since this multipartite ranking is based on the class information, it can generate a coherent ranking of features, for both of the training and test sets. This also leads to an efficient spread of responses, and effective generalization for deep convolutional neural networks.

In summary, the proposed multipartite pooling method has several advantages. It considers the distribution of each class and calculates the rank of individual features. Due to the data-driven process of scoring, the performance gap between training-test errors, is considerably closer. It generates superior performance on standard benchmark datasets, in comparison with the average, maximum and stochastic pooling schemes, when identical evaluation protocols are applied. The conducted experiments on various benchmarks, confirm that the proposed strategy of multipartite pooling, consistently improves the performance of deep convolutional networks, by using better model generalization for the test-time data.

\newpage

\section{Formulation}

\begin{figure}[!t]
\begin{center}
\includegraphics[width=0.45\linewidth]{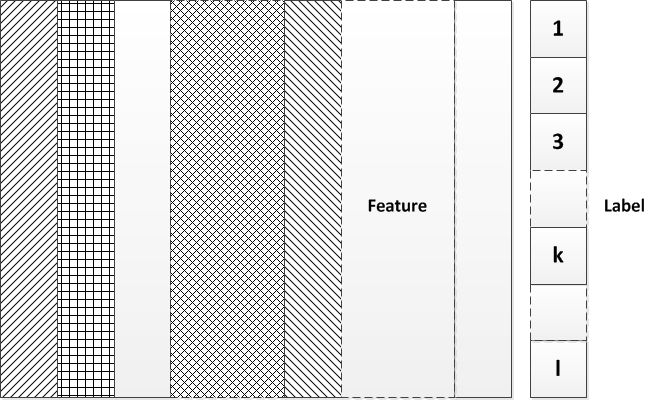}
\hspace*{0.05\linewidth}
\includegraphics[width=0.45\linewidth]{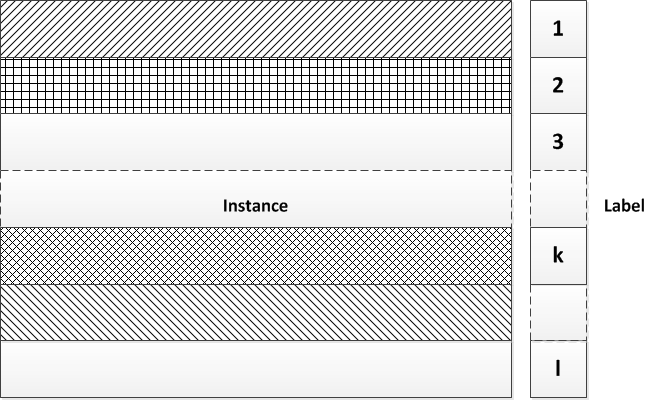}
\end{center}
\caption[Feature vs instance ranking.]{Feature vs instance ranking. A set of features (columns) and instances (rows) are assigned to $|\mathcal{L}|$ different labels. They are ranked upon their separability, represented by different line patterns and are scored. These scores are used for selecting the best features or instances. To employ either of them for convolutional pooling, the labels must be known. The problem is that, classic feature-instance ranking methods are specific to the training-time data, and there is no way to exploit them for pooling of the test-time data. To solve this inconsistency, the notion of labels is mapped to the test data and then, instance ranking strategies are applied to the pooling layers. This is accomplished by the supervised projection.}
\label{fig:feature-instance}
\end{figure}

This section begins with multipartite ranking and moves towards the multipartite pooling. The multipartite ranking means scoring of each representation in the feature set, with respect to the distinctive information. Instances with higher scores are expected to be more informative, and hence, receive higher ranks. The intuition of multipartite pooling is about picking the activation instances with the higher scores (ranks) in a pooling window, to achieve better activations in the pooling layer. A graphical interpretation of feature vs instance ranking is depicted in Figure~\ref{fig:feature-instance}, where columns represent the activations.

For a two-class regime, the criterion to calculate the significance of each feature, can be selected from statistical measures, such as, absolute value two-sample t-test with pooled variance estimate~\cite{jain2003local}; relative entropy or Kullback-Leibler distance~\cite{hershey2007approximating}; minimum attainable classification error or Chernoff bound~\cite{nussbaum2009chernoff}; area between the empirical Receiver Operating Characteristic (ROC) curve and the random classifier slope~\cite{fawcett2006introduction}; and absolute value of the standardized u-statistic of a two-sample unpaired Wilcoxon test or Mann-Whitney test~\cite{bohn1994effect}.

Suppose that a set of instances $\mathcal{X}=\{X_1,\dots,X_{|\mathcal{X}|}\}$ are assigned to a set pf predefined label $\mathcal{L}=\{L_1,\dots,L_{|\mathcal{L}|}\}$, such that, $\mathcal{X}$ is a matrix with $|\mathcal{X}|$ instances (rows). The aim is to rank the features (columns), using an independent evaluation criterion. This criterion is a distance that measures the significance of an instance, for imposing higher class distinction in the set $\mathcal{X}$. The absolute value of the criterion for a bipartite ranking scenario, with only two valid labels $\{L_1,L_2\}$, is defined as,

\begin{equation}
\mathcal{C}_B(\mathcal{X}) = KL_{12}(X_1,\dots,X_{|\mathcal{X}|})
\label{eq:object:01}
\end{equation}

\newpage

Here, $KL$ is the Kullback-Leibler divergence and $\mathcal{C}_B(\mathcal{X})$ is the binary criterion measured for each feature (column) of the set $\mathcal{X}$. This equation can be extended to the summation of binary criteria, where each labels is considered as primary label (foreground) and the rest are merged as secondary labels (background). 

The overall criterion of the multipartite case, with multiple labels $\mathcal{L}$, can be formulated as,

\begin{equation}
\mathcal{C}_M(\mathcal{X}) = \sum_{i=1,\;j\neq{i}}^{i={|\mathcal{L}|}} KL_{ij}(X_1,\dots,X_{|\mathcal{X}|})
\label{eq:object:02}
\end{equation}

\noindent where $KL_{ij}$ is the cumulative Kullback-Leibler distance of label $L_i$ to the rest of the labels of set $\mathcal{L}$, which are $\forall\;L_j\in\mathcal{L}-L_i$. It is clear that, higher values of $\mathcal{C}_M(\mathcal{X})$ for a feature, means better class separability. A high-ranked representation is beneficial to any classifier, because there are better distinctions between classes in the set $\mathcal{X}$.

It is possible to employ the above formulation for instance ranking. In other words, instances (rows) are ranked instead of features (columns), which is required for pooling operation, where high-ranked instances are selected as the representations for each convolutional filters. In contrast with feature ranking, the rows of set $\mathcal{X}$, which correspond to convolutional representations, are ranked in the pooling layers. 

To connect the features into instances, a projection from the feature space into a new instance space, spanned by the number of classes in $\mathcal{X}$, is employed. In this space, a new set $\mathcal{P}$ is created by multiplying the feature set $\mathcal{X}$ with a projection matrix $\mathbf{A}$, such that,

\begin{equation}
\mathcal{C}_M(\mathcal{P}) = \sum_{i=1,\;j\neq{i}}^{{i={|\mathcal{L}|}}} KL_{ij}(X_1{\mathbf{A}},\dots,X_{|\mathcal{X}|}{\mathbf{A}})
\label{eq:object:03}
\end{equation}

\noindent where the set $\mathcal{P}$ is a matrix with $|\mathcal{X}|$ instance (rows) and $|\mathcal{L}|$ features (columns). The projection matrix $\mathbf{A}$ enables the same ranking strategies for features of the set $\mathcal{X}$, to be applied to the instances of the $\mathcal{P}$, so that, the highly-ranked activations are selected.

\begin{figure}[!t]
\begin{center}
\includegraphics[width=0.45\linewidth]{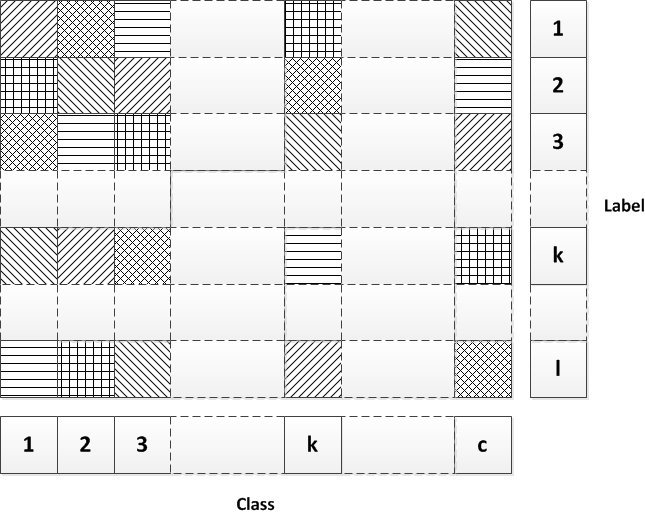}
\hspace*{0.05\linewidth}
\includegraphics[width=0.45\linewidth]{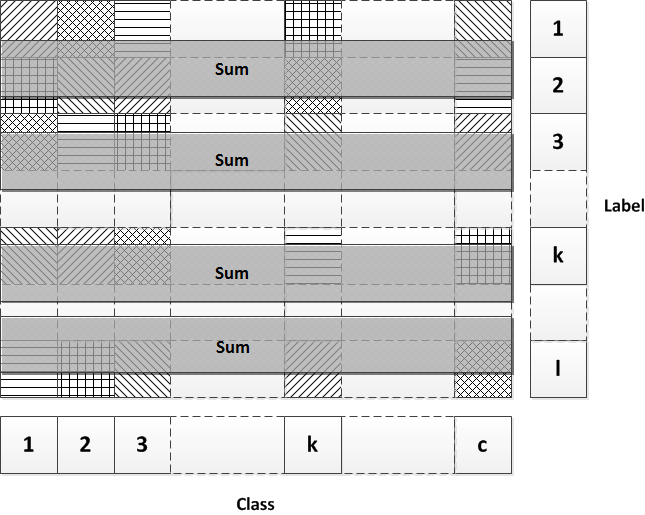}
\end{center}
\caption[Multipartite ranking.]{Multipartite ranking. The projected set $\mathcal{P}$ is ranked and the scores are aggregated to compute overall criterion $\mathcal{C}_M(\mathcal{P})$. Since the columns represent classes inside the feature set $\mathcal{X}$, bipartite rank of each columns, is calculated with respect to the rest of columns. This generates $c=|\mathcal{L}|$ different scores, represented by different line patterns, for each of the $|\mathcal{P}|$ instances. By sliding an accumulative bar, represented by grey rectangle, the overall score is computed for each instances. These overall scores are used to rank and pool the most informative instances, which are activations of the pooling layers.}
\label{fig:ranking}
\end{figure}

\subsection{Supervised Projection}
\label{sec:projection}

To formulate the above projection, the matrix $\mathbf{A}$ can be considered as a mapping, which tries to project $\mathcal{X}$ into a space with $c=|\mathcal{L}|$ dimensions. The projection matrix $\mathbf{A}$ is determined to minimize the Fisher criterion given by,

\begin{equation}
\mathcal{J}(\mathbf{A})=tr\bigg[(\mathbf{A}\mathbf{S_w}\mathbf{A}^T)(\mathbf{A}\mathbf{S_b}\mathbf{A}^T)^{-1}\bigg]
\label{eq:object:04}
\end{equation}

\noindent that $\mathit{tr(.)}$ is the diagonal summation operator. The within ($\mathbf{S_w}$) and between ($\mathbf{S_b}$) class scatterings are defined as,

\begin{equation}
\mathbf{S_w} = \sum_{j=1}^c\sum_{x_i \in\mathbf{C}_j}(x_i - \mu_j)(x_i - \mu_j)^T
\label{eq:object:05}
\end{equation}

\begin{equation}
\mathbf{S_b} = \sum_{j=1}^c(\mu_j-\bar{\mu})(\mu_j-\bar{\mu})^T
\label{eq:object:06}
\end{equation}

\noindent where $c$, $\mu_j$ and $\bar{\mu}$ are number of classes, mean over class $\mathbf{C}_j$ and mean over the set $\mathcal{X}$. The matrix $\mathbf{S_w}$ can be regarded as average class-specific covariance, whereas $\mathbf{S_b}$ can be viewed as the mean distance between all different classes. The purpose of Equation~\ref{eq:object:04} is to maximize the between-class scattering, while preserving within-class dispersion. The solution can be retrieved from a generalized eigenvalue problem $\mathbf{S_b}\mathbf{A}= \lambda\mathbf{S_w}\mathbf{A}$. For $c=|\mathcal{L}|$ classes, the projection matrix $\mathbf{A}$ builds upon eigenvectors corresponding to the largest $c$ eigenvalues of $\mathbf{S^{-1}_w}\mathbf{S_b}$~\cite{bishop2006pattern}. 

\begin{algorithm}[!t]
	\caption{Supervised Projection}
	\label{alg:projection}
	\begin{algorithmic}
		\STATE
		\STATE {\bfseries Input:} feature set $\mathcal{X}$
		\STATE {\bfseries Output:} projection matrix $\mathbf{A}$
		\STATE
		\STATE 1: Compute $\mathbf{S_w}$~(Equation~\ref{eq:object:05}) and $\mathbf{S_b}$~(Equation~\ref{eq:object:06})
		\STATE 2: Set $\mathbf{A^{(0)}}$ as largest eigenvalues of $\mathbf{S_w^{-1}}\mathbf{S_b}$
		\STATE 3: Minimize Equation~\ref{eq:object:07} by using Equation~\ref{eq:object:10}
		\STATE
	\end{algorithmic}
\end{algorithm}

To yield better distinction, the ratio of between and within class scatterings (quotient-of-trace) is minimized~\cite{cunningham2015linear}, by imposing orthogonality to the following cost function,

\begin{equation}
\mathcal{Q(\mathbf{A})}=tr({\mathbf{A}\mathbf{S_w}\mathbf{A^T}})\big[tr\big({\mathbf{A}\mathbf{S_b}\mathbf{A^T}})\big]^{-1}+\lambda\lVert\mathbf{I}-{\mathbf{A}\mathbf{A^T}}\rVert_2
\label{eq:object:07}
\end{equation}

The first part of function $\mathcal{Q(\mathbf{A})}$ defines a form of Fisher criterion that aims for making the highest possible separability among classes. The second term is a regularization term imposing orthogonality into the projection matrix $\mathbf{A}$. Looking back at Equation~\ref{eq:object:04}, it can be seen that the set of eigenvectors corresponding to the largest $c$ eigenvalues of $\mathbf{S_w^{-1}}\mathbf{S_b}$ is a solution for the above optimization problem. This can be taken as an initial projection matrix $\mathbf{A^{(0)}}$. 

Now, it is possible to minimize $\mathcal{Q(\mathbf{A})}$ by using stochastic gradient descent. Here, $\mathbf{A^{(0)}}$ is employed as an initialization point, because conventional Fisher criterion is the trace-of-quotient, which can be solved by generalized eigenvalue method. Equation~\ref{eq:object:07} is the quotient-of-trace, that requires a different solution~\cite{cunningham2015linear}.

\newpage

To work out the closed form derivatives of Equation~ \ref{eq:object:07}, suppose that $\mathcal{Q(\mathbf{A})}$ is composed of $\mathcal{Q}_{1}(\mathbf{A})$ and $\mathcal{Q}_{2}(\mathbf{A})$ as follows,

\begin{equation}
\mathcal{Q}_{1}(\mathbf{A}) = tr({\mathbf{A}\mathbf{S_w}\mathbf{A^T}})\big[tr({\mathbf{A}\mathbf{S_b}\mathbf{A^T}})\big]^{-1}
\label{eq:object:08}
\end{equation}

\begin{equation}
\mathcal{Q}_{2}(\mathbf{A}) = \lambda\lVert\mathbf{I}-{\mathbf{A}\mathbf{A^T}}\rVert_2
\label{eq:object:09}
\end{equation}

According to matrix calculus~\cite{petersen2008matrix},

\begin{eqnarray}
\frac{\partial{\mathcal{Q}}}{\partial{\mathbf{A}}} & = & \frac{{tr({\mathbf{A}\mathbf{S_w}\mathbf{A^T}})}}{{\big[tr({\mathbf{A}\mathbf{S_b}\mathbf{A^T}})}\big]^2}
\big(\mathbf{S^T_b}+\mathbf{S_b}\big)\mathbf{A^T}
\nonumber\\
&-& \frac{1}{{tr({\mathbf{A}\mathbf{S_b}\mathbf{A^T}})}}
\big(\mathbf{S^T_w}+\mathbf{S_w}\big)\mathbf{A^T}
\nonumber\\
&-& \frac{2\lambda}{\lVert\mathbf{I}-{\mathbf{A}\mathbf{A^T}}\rVert_2}
\;\mathbf{A^T}(\mathbf{I}-{\mathbf{A}\mathbf{A^T}})
\label{eq:object:10}
\end{eqnarray}

The computation of $\mathbf{A}$ is summarized in Algorithm~\ref{alg:projection}. For implementation purposes, the built-in function of Matlab optimization toolbox~\cite{coleman1996reflective} is employed.

\subsection{Multipartite Ranking}
\label{sub:multipartite-ranking}

\begin{algorithm}[!t]
\caption{Multipartite Ranking}
\label{alg:ranking}
\begin{algorithmic}
\STATE
\STATE {\bfseries Input:} feature set $\mathcal{X}$, label set $\mathcal{L}$
\STATE {\bfseries Output:} overall criterion $\mathcal{C}_M(\mathcal{P})$
\STATE
\STATE 1: Compute the projection matrix $\mathbf{A}$~(Algorithm~\ref{alg:projection})
\STATE 2: Calculate the projected set $\mathcal{P}=\mathcal{X}\mathbf{A}$
\STATE
\FOR{$i=1$ {\bfseries to} $|\mathcal{L}|$}
\STATE 3: Split $\mathcal{P}$ between labels $\{L_i,L_j\}$, when $L_j\in\mathcal{L}-L_i$
\STATE 4: Calculate $KL_{ij}(\mathcal{P})$ (Equation~~\ref{eq:object:03})
\ENDFOR
\STATE
\STATE 5: Set $\mathcal{C}_M(P)=\sum KL_{ij}(\mathcal{P})$
\STATE
\end{algorithmic}
\end{algorithm}

Drawing upon the above information, it is possible to put forward the proposed multipartite ranking scheme. Using the instance ranking strategy, one can take the feature set $\mathcal{X}$, deploy the supervised projection $\mathbf{A}$ (Algorithm~\ref{alg:projection}) to produce the projected set $\mathcal{P}$, and calculate the cumulative Kullback-Leibler distance (Equation~\ref{eq:object:03}), as the ranking scores, for each instance of the projected set $\mathcal{P}$. 

Since the number of instances in $\mathcal{P}$ is equal to the number of instances in $\mathcal{X}$, and these two matrices are related linearly through $\mathbf{A}$ via $\mathcal{P}=\mathcal{X}\mathbf{A}$, the overall criterion $\mathcal{C}_M(\mathcal{P})$ is sorted to rank the instances of the set $\mathcal{X}$, in regard to their class separability. Algorithm~\ref{alg:ranking} represents the multipartite ranking method. The process of the multipartite ranking is also visualized in Figure~\ref{fig:ranking}. Each column of the set $\mathcal{P}$ represents a specific class of the set $\mathcal{X}$ and hence, the Kullback-Leibler binary scoring scheme (one-versus-all) is employed to set a criterion measure for each of its individual instances (rows). After it has been applied to all the columns, it starts to scan rows and accumulate scores, resulting in the overall criteria, $\mathcal{C}_M(\mathcal{P})$. This is then used to rank the instances of the projected set $\mathcal{P}$.

\newpage

The reason for projecting to the space spanned by the number of classes, is to use the above, one-versus-all strategy. The bipartite ranking by Kullback-Leibler divergence requires that, one main class is selected as the foreground label, while those remaining are used as background labels. It gives a measure of how the foreground is separated from the background data. It is necessary to use statistics to ensure that the cumulative criterion, $\mathcal{C}_M(\mathcal{P})$ is a true representation of the all instances, contained within the set $\mathcal{X}$.

When $\mathcal{X}$ is projected to lower dimensions than the number of available classes, the result is that, some of the classes are missed. In contrast, projection of $\mathcal{X}$ to higher dimensions than the number of classes, leads to partitioning of some classes to pseudo labels, that are not queried during the test phase. Either way, the generated scores are not reliable for the sake of pooling, because they are not derived from the actual distribution of the classes.

\subsection{Multipartite Pooling}
\label{multipartite pooling}

The above multipartite ranking strategy can be employed for the pooling. In general, a deep convolutional neural network consists of consecutive convolution and pooling layers. The convolutional layers extract common patterns within local patches. Then, a nonlinear elementwise operator is applied to the convolutional features, and the resulting activations are passed to the pooling layers. These activations are less sensitive to the precise locations of structures within the data, than the primary features. Therefore, the consecutive convolutional layers can extract features, that are not susceptible to spatial transformations or distortions of the input~\cite{zeiler2013stochastic}.

Suppose that a stack of convolutional features $\mathcal{S}=\{S_1,\dots,S_{|\mathcal{S}|}\}$ passes through the pooling layer. The matrix $\mathcal{S}$ is an array of dimensions $h\times{w}\times{d}\times{n}$ where $h$ and $w$ are height and width of samples, $d$ is depth of stack (number of filters), such that $S_i\in\mathbb{R}^{h\times{w}\times{d}}$ creates a three-dimensional vector, and $n$ is number of samples in the stack ($n=|\mathcal{S}|$). 

\begin{algorithm}[!t]
\caption{Multipartite Pooling}
\label{alg:pooling}
\begin{algorithmic}
\STATE
\STATE {\bfseries Input:} convolutional feature stack $\mathcal{S}\in\mathbb{R}^{h\times{w}\times{d}\times{n}}$
\STATE {\bfseries Output:} overall criterion $\mathcal{C}_M(\mathcal{S})$
\STATE
\STATE 1: Reshape each $S_i\in\mathbb{R}^{h\times{w}\times{d}}$ to $S_i\in\mathbb{R}^{hw\times{d}}$
\STATE 2: Concatenate all $S_i$ columns to give $\mathcal{X}\in\mathbb{R}^{hwn\times{d}}$
\STATE 3: Calculate $\mathcal{C}_M(\mathcal{P})\in\mathbb{R}^{hwn\times{1}}$ by Algorithm~\ref{alg:ranking}
\STATE 4: Partition $\mathcal{C}_M(\mathcal{P})$ to give $\mathcal{C}_{M}(\mathcal{S})\in\mathbb{R}^{h\times{w}\times{n}}$
\STATE 5: Pool the activations based on $\mathcal{C}_{M_i}\in\mathbb{R}^{h\times{w}}$ for all $i\in[1:n]$
\STATE
\end{algorithmic}
\end{algorithm}

The standard pooling methods either retain the maximum or average value, over the pooling region per channel. The multipartite pooling method begins with the reshaping of feature stack $\mathcal{S}$ to form $n$ two-dimensional $X_i\in\mathbb{R}^{hw\times{d}}$. The elements of set $\mathcal{X}=\{X_1,\dots,X_{|\mathcal{X}|}\}$ are concatenated such that $|\mathcal{X}|=n$ and $\mathcal{X}\in\mathbb{R}^{hwn\times{d}}$ is a two-dimensional matrix. Now, $\mathcal{X}$ is ready to deploy Algorithm~\ref{alg:ranking} and compute the overall criterion $\mathcal{C}_M(\mathcal{P})\in\mathbb{R}^{hwn\times{1}}$. Partitioning to $n$ and reshaping into $h\times{w}$ windows, give $\mathcal{C}_M(\mathcal{S})=\{\mathcal{C}_{M_1},\dots,\mathcal{C}_{M_{|\mathcal{S}|}}\}$, that $\mathcal{C}_{M_i}\in\mathbb{R}^{h\times{w}}$ provides the rank of each pixels of $X_i$. To apply the pooling, a sliding window goes through each region and picks the representation with the greatest $\mathcal{C}_M(\mathcal{S})$. These are the activations with the best separation among available classes.

As a numerical example, consider MNIST dataset with $|\mathcal{L}|=10$ classes. The first pooling layer is fed by a stack $\mathcal{S}$, consisting of convolution responses of $d=20$ filters, with $n=100$ frames, of size $w=24$ by $h=24$ pixels. First, $\mathcal{S}$ is reshaped to $100$ samples of size $576\times{20}$ pixels, form the set $\mathcal{X}$, which will be concatenated as a $57600\times{20}$ array. Second, the projection matrix $\mathbf{A}\in\mathbb{R}^{20\times{10}}$ is calculated to project $\mathcal{X}$. Third, $\mathcal{C}_M(\mathcal{P})\in\mathbb{R}^{57600\times{1}}$ is computed and partitioned into $100$ criterion measures $\mathcal{C}_{M_i}$, of size $24\times{24}$ pixels. For the pooling of $\mathcal{S}$, a $2\times{2}$ window moves along each frame and picks the top-score pixels. The output is a set of $100$ features of size $12\times{12}$ pixels for each of $20$ convolutional filters.

\newpage

\section{Experiments}
\label{experiments}

For evaluation purposes, the multipartite pooling is compared with the popular maximum, average and stochastic poolings. A standard experimental setup~\cite{zeiler2013stochastic} is followed to apply the multipartite pooling for MNIST, CIFAR and Street View House Numbers (SVHN) datasets. The results show that, when multipartite pooling is employed to pool convolutional features, lower test error rates than other pooling strategies, are achieved. For implementation, the library provided by the Oxford Visual Geometry Group~\cite{vedaldi08vlfeat} is used.

\subsection{Datasets}
\label{datasets}

\begin{figure}[!t]
\begin{center}
\includegraphics[width=1.0\textwidth]{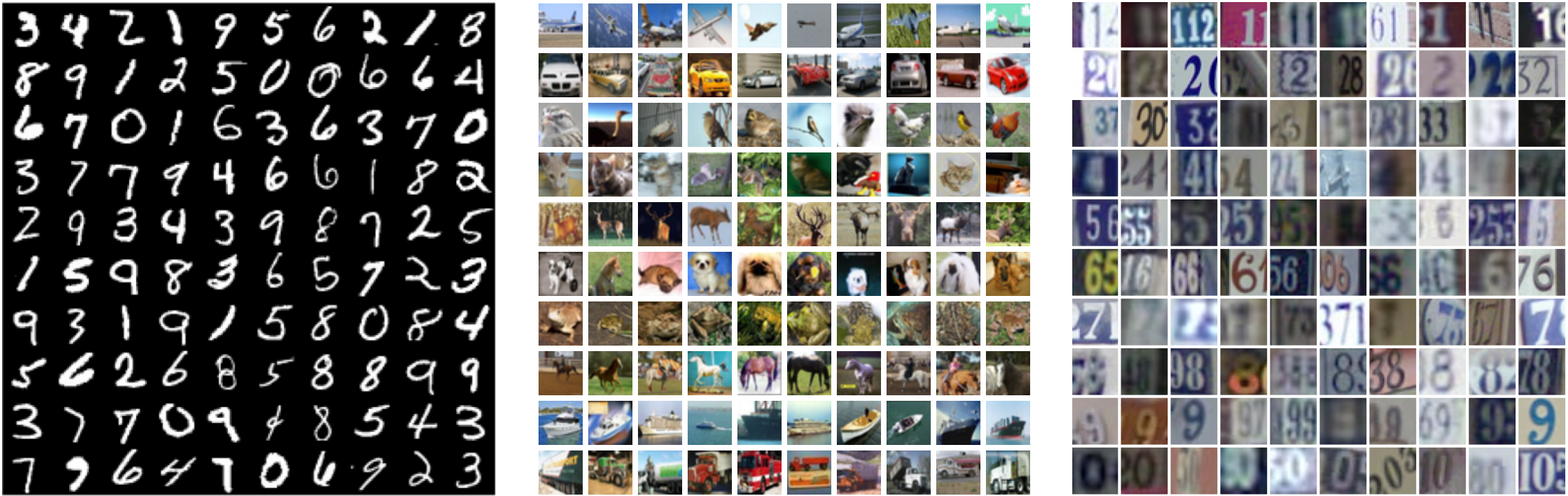}
\end{center}
\caption[Examples of images from MNIST, CIFAR and SVHN.]{Examples of images from MNIST, CIFAR and SVHN Datasets.}
\label{fig:datasets}
\end{figure}

The MNIST dataset~\cite{lecun1998gradient}, contains $60,000$ training examples, and $10,000$ test samples, normalized to $20\times{20}$ pixels, centred by centre of mass in $28\times{28}$ pixels, and sheared by horizontally shifting, such that, the principal axis is vertical. The foreground pixels were set to one, and the background to zero. 

The CIFAR dataset~\cite{krizhevsky2009learning}, includes two subsets. The first subset, CIFAR-10 consists of $10$ classes of objects with $6,000$ images per class. The classes are airplane, automobile, bird, cat, deer, dog, frog, horse, ship, and truck. It was divided into $5,000$ randomly selected images per class as training set, and the rest served as test samples. The second subset, CIFAR-100 has $600$ images for each of $100$ classes. These classes also come in $20$ super-classes, each consisting of five classes.

The SVHN dataset~\cite{netzer2011reading}, was extracted from a large number of Google Street View images by automated algorithms and the Amazon Mechanical Turk (AMT). It consists of over $600,000$ labelled characters in full numbers and MNIST-like cropped digits in $32\times{32}$ pixels. Three subsets are available containing $73,257$ digits for training, $26,032$ for testing and $531,131$ extra samples.

\subsection{Results \& Discussion}
\label{results}

The following tables represent the evaluation results for four different schemes, including max, average, stochastic, and multipartite poolings, in terms of the training and test errors. To gain an insight into added computational load, one should recall that the only intensive calculations are related to working out the supervised projection, at the training phase. This procedure (Algorithm~\ref{alg:projection}), depends on the number of activations ($h\times{w}\times{n}$ instances) and convolutional filters ($d$ dimensions), which are employed in Algorithm~\ref{alg:pooling}. For testing, one just multiplies and no computational cost is involved. For example, the number of samples, trained in an identical architecture for MNIST dataset, are $285$, and $195$ samples per second for the max and multipartite pooling strategies, respectively.

The classification performance on MNIST dataset is reported in Table~\ref{tab:mnist}. It can be seen that, max pooling gives the best training performance, but its test error is larger than the stochastic and multipartite pooling. In other words, it may overfit for MNIST, although its test performance is higher than the average pooling. The multipartite pooling performs better than all other schemes, despite greater training error, compared to max and stochastic pooling. 

\begin{table}[t]
\begin{center}
\begin{tabular}{|l||c|c|}
\hline
Pooling & Train (\%) & Test (\%)\\
\hline\hline
Average & 0.57 & 0.83 \\
Max & \textbf{0.04} & 0.55 \\
Stochastic & 0.33 & 0.47 \\
Multipartite (Proposed) & 0.38 & \textbf{0.41} \\
\hline
\end{tabular}
\end{center}
\caption[Classification errors for different pooling strategies on MNIST.]{Classification errors for different pooling strategies on MNIST dataset. The multipartite pooling approach provides the lowest test error, in spite of high training error. This is due to better generalization of the proposed pooling, compared to the other methods.}
\label{tab:mnist}
\begin{center}
\begin{tabular}{|l||c|c|}
\hline
Pooling & Train (\%) & Test (\%)\\
\hline\hline
Average & 1.92 & 19.24 \\
Max & \textbf{0.0} & 19.40 \\
Stochastic & 3.40 & 15.13 \\
Multipartite (Proposed) & 12.63 & \textbf{13.45} \\
\hline
\end{tabular}
\end{center}
\vfill
\begin{center}
\begin{tabular}{|l||c|c|}
\hline
Pooling & Train (\%) & Test (\%)\\
\hline\hline
Average & 11.20 & 47.77 \\
Max & \textbf{0.17} & 50.90 \\
Stochastic & 21.22 & 42.51 \\
Multipartite (Proposed) & 36.32 & \textbf{40.81} \\
\hline
\end{tabular}
\end{center}
\caption[Classification errors for different pooling strategies on CIFAR.]{Classification errors for different pooling strategies on CIFAR-10 and CIFAR-100 datasets. The multipartite pooling outperforms other pooling schemes on test errors, but it is behind on training errors. The close gap between training and test errors, leads to better classification performance for the proposed pooling strategy.}
\label{tab:cifar}
\begin{center}
\begin{tabular}{|l||c|c|}
\hline
Pooling & Train (\%) & Test (\%)\\
\hline\hline
Average & 1.65 & 3.72 \\
Max & \textbf{0.13} & 3.81 \\
Stochastic & 1.41 & 2.80 \\
Multipartite (Proposed) & 2.09 & \textbf{2.15} \\
\hline
\end{tabular}
\end{center}
\caption[Classification errors for different pooling strategies on SVHN.]{Classification errors for different pooling strategies on SVHN dataset. The multipartite pooling scheme gives the best performance on the test, that is closely followed by the training error.}
\label{tab:svhn}
\end{table}

\newpage

The multipartite pooling is more competent because it draws upon the statistics of training-test data for pooling. It is in contrast to picking the maximum response (max pooling); smoothing the activations (average pooling); and random selection (stochastic pooling); which disregard the data distribution. In a pooling layer of any deep learning architecture, aggregation of the best available features is critical to infer complicated contexts, such as, shape derived from primitives of edge and corners. The proposed pooling learns how to pick the best informative activations from the training set, and then, employs this knowledge at the test phase. As a result, the performance consistently improves in all the experiments.

\begin{figure}[p]
\begin{center}
\includegraphics[width=0.6\textwidth]{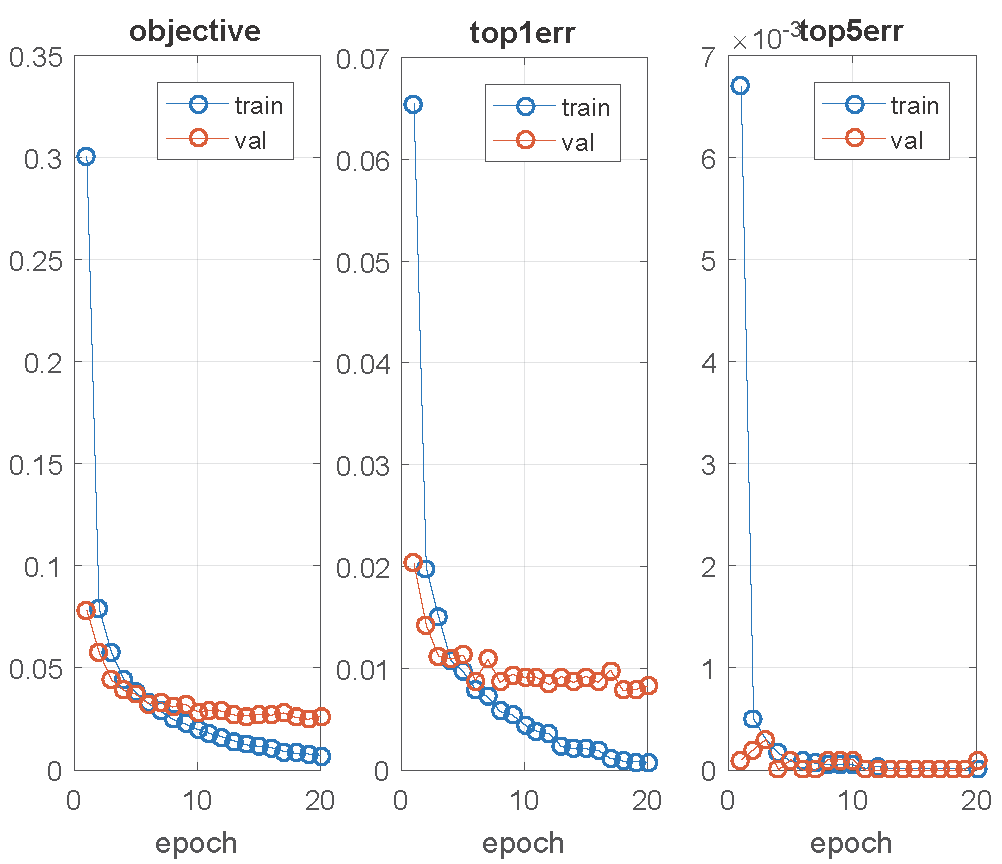}
\end{center}
\caption[Max pooling for MNIST.]{Max pooling for MNIST. The first graph represents the loss function (objective) for both training (train) and test (val) on MNIST dataset. The other graphs correspond to the top 1 (top1err) and the top 5 (top5err) errors.}
\label{fig:mnist-base}
\vfill
\begin{center}
\includegraphics[width=0.6\textwidth]{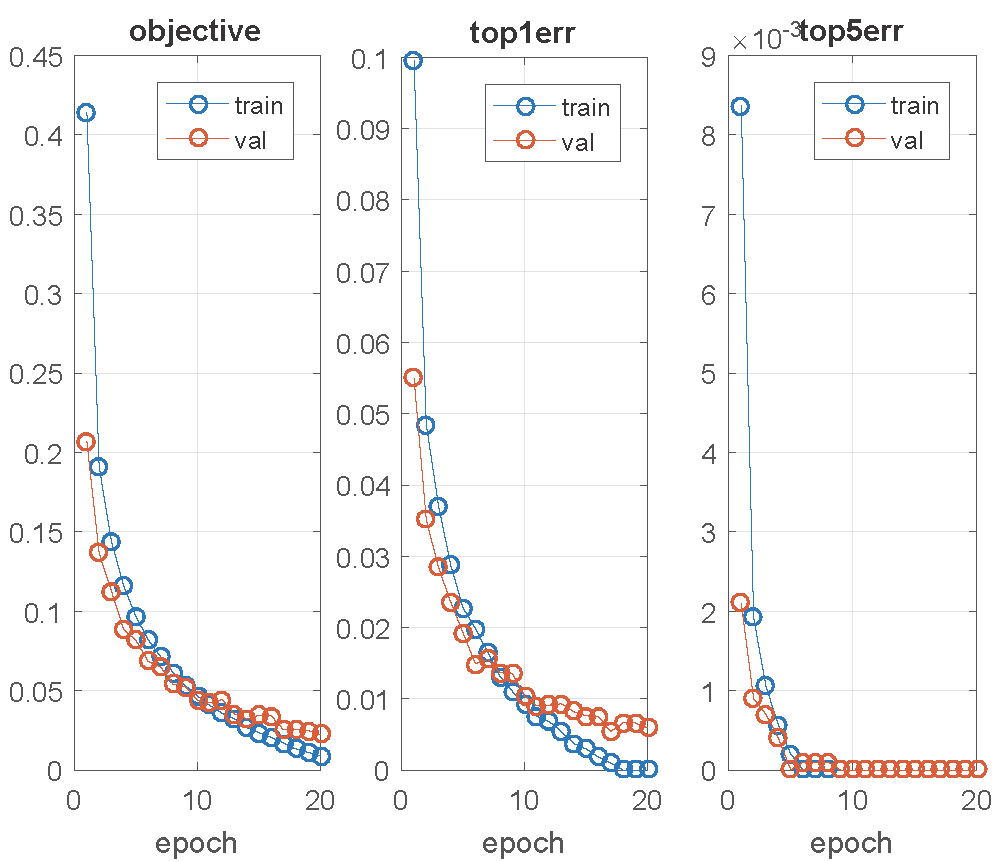}
\end{center}
\caption[Multipartite pooling for MNIST.]{Multipartite pooling for MNIST. In comparison with the max pooling (Figure~\ref{fig:mnist-base}), the test loss and errors (val) are reduced by applying the multipartite pooling technique. Note that the training and test performances get closer to each other, indicating better generalization of the trained network.}
\label{fig:mnist-pool}
\end{figure}

Figures~\ref{fig:mnist-base} and~\ref{fig:mnist-pool} show the training-test performances of MNIST dataset for $20$ epochs. Except for the early epochs, they are quite close to each other in the multipartite pooling regime. The reason is that, both of the training-test poolings are connected with a common factor; the projection matrix $\mathbf{A}$. This is trained with the training set and is deployed by multipartite pooling on the test set, to pick the most informative activations. Since the same criterion (Kullback-Leibler) is employed to rank the projected activations for training-test, and they are mapped with the same projection matrix $\mathbf{A}$, it is expected that the trained network will demonstrate better generalization than alternative pooling schemes, where the training-test poolings are statistically disconnected. The graphs show that the multipartite pooling generalizes considerably better.

\newpage

Tables~\ref{tab:cifar} provides the performance metrics for CIFAR datasets. It is apparent that the multipartite pooling outperforms other approaches on the test performance. It also prevents the model from overfitting, in contrast to the max pooling. In the proposed pooling method, better generalization also contributes to another advantage; preventing under-overfitting. As mentioned before, pooling at the test phase is linked to the training phase by the projection. This ensures that the test performance follows the training closely and hence, it is less likely to end up under-overfitting.

\begin{figure}[p]
\begin{center}
\includegraphics[width=0.6\textwidth]{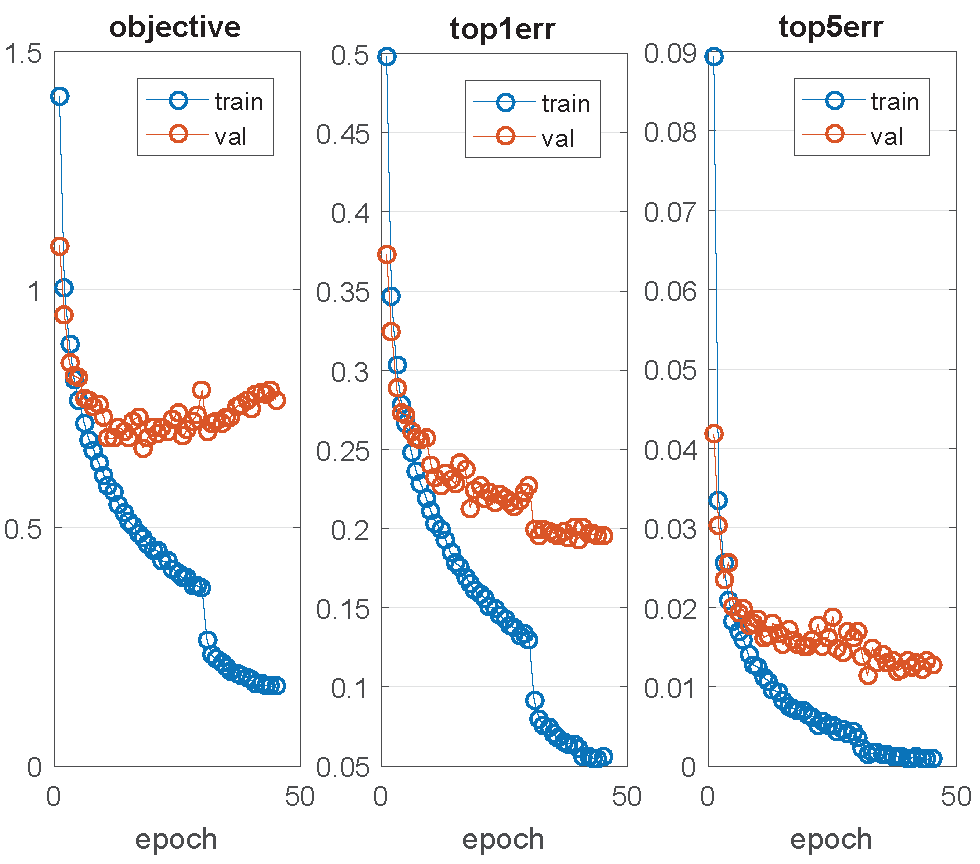}
\end{center}
\caption[Max pooling for CIFAR-10.]{Max pooling for CIFAR-10. Compared to MNIST, this results in greater loss and errors on CIFAR-10, due to the variety in samples and tasks (character vs object recognition). The gap between the training (train) and test (val) errors, is considerably wider for CIFAR-10.}
\label{fig:cifar10-base}
\vfill
\begin{center}
\includegraphics[width=0.6\textwidth]{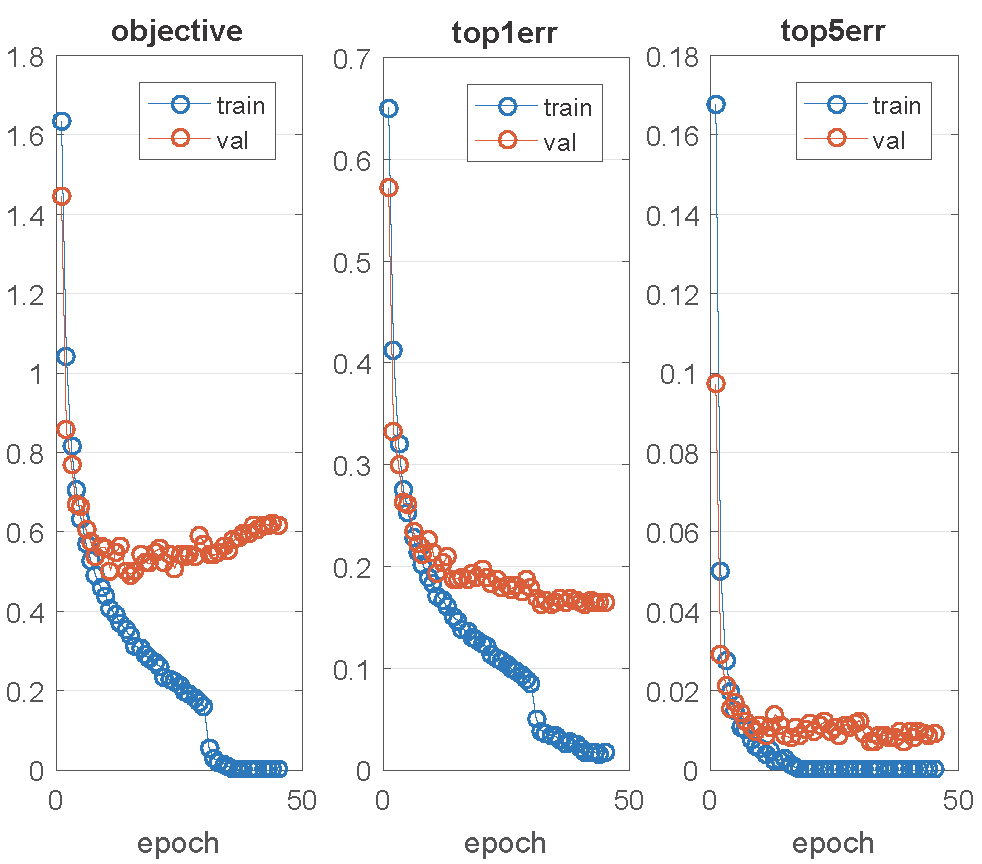}
\end{center}
\caption[Multipartite pooling for CIFAR-10.]{Multipartite pooling for CIFAR-10. Besides smaller losses and errors, with respect to the max pooling (Figure~\ref{fig:cifar10-base}), the training (train) and test (val) performances indicate closer gaps, due to better generalization of the model, using by multipartite pooling.}
\label{fig:cifar10-pool}
\end{figure}

One striking observation is that the gap between the training-test errors is wider for CIFAR-100 compared to CIFAR-10. This relates to the number of classes in each subsets of CIFAR. Since CIFAR-100 has more classes, it is more difficult to impose separability, hence, the difference between the training and test performances will increase. Figures~\ref{fig:cifar10-base} and \ref{fig:cifar10-pool} depicts the errors of CIFAR-10 dataset for $50$ epochs. It is clear that employing of the max pooling results in a huge gap between the training-test performances due to the overfitting.

Finally, the evaluation outcomes for SVHN dataset are presented in Table~\ref{tab:svhn}. Again, the multipartite pooling does a better job on the test. The error is close to the stochastic pooling method. This implies that when the number of samples in a dataset increases greatly, the multipartite ranking scores lean towards the probability distribution, generated by the stochastic pooling. Here, any infinitesimal numerical errors may also lead to an inaccurate pooling, which may, in return, degrade the informative activations of a layer for both of the pooling methods. Since the multipartite pooling takes a deterministic approach, the effect of numerical inconsistencies, is considerably smaller than stochastic pooling, which randomly picks activations on a multinomial distribution~\cite{zeiler2013stochastic}).

Overall, the employment of multipartite ranking for the purpose of pooling in deep convolutional neural networks, produces superior results compared to all the other strategies, tested in the experiments. It is robust to overfitting and shows better generalization characteristics, by connecting the training and test statistics.

\section{Conclusion}
\label{conclusion}

We introduce a novel pooling strategy called multipartite pooling that is based on multipartite ranking of the features in pooling layers of deep convolutional neural networks. This pooling scheme projects the features to a new space and then, score them by an accumulative bipartite ranking approach. These scores can be used to pick the highly informative activations in the pooling layers of any deep convolutional neural networks. We conduct our experiments on four publicly available datasets (MNIST, CIFAR-10, CIFAR-100, and SVHN) and report the errors of four different pooling schemes (maximum, average, stochastic, and multipartite). The results show that our proposed multipartite pooling method outperforms all other pooling strategies in all datasets and provides a more efficient generalization for the deep learning architectures.

\newpage

\bibliographystyle{nips}
\bibliography{nips}

\end{document}